\documentclass[10pt,twocolumn,a4paper]{esaAI}

\title{Training Datasets Generation for Machine Learning: 
Application to Vision Based Navigation}

\def\authorEmail{surrender.software@airbus.com}

\author[1]{Jérémy Lebreton\thanks{Corresponding author. E-Mail: \authorEmail}}
\author[2]{Ingo Ahrns}
\author[1]{Roland Brochard}
\author[2]{Christoph Haskamp}
\author[3]{Hans Krüger}
\author[1]{Matthieu Le\,Goff}
\author[1]{Nicolas Menga}
\author[1]{Nicolas Ollagnier}
\author[2]{Ralf Regele}
\author[4]{Francesco Capolupo}
\author[4]{Massimo Casasco}
\affil[1]{Airbus Defence and Space, Toulouse}
\affil[2]{Airbus Defence and Space, Bremen}
\affil[3]{DLR Institute for Space Systems}
\affil[4]{European Space Agency, Noordwijk}

\begin{document}

% Creates the title and author list automatically for you!
\makeCustomtitle

% EDIT HERE
% Please add your abstract here, i.e., \begin{abstract}<Your abstract text>\end{abstract}
\begin{abstract}
Vision Based Navigation consists in utilizing cameras as precision sensors for GNC after extracting information from images. To enable the adoption of machine learning for space applications, one of obstacles is the demonstration that available training datasets are adequate to validate the algorithms. 
The objective of the study is to generate datasets of images and metadata suitable for training machine learning algorithms. 
Two use cases were selected and a robust methodology was developed to validate the datasets including the ground truth. The first use case is in-orbit rendezvous with a man-made object: a mockup of satellite ENVISAT. The second use case is a Lunar landing scenario. Datasets were produced from archival datasets (Chang’e\,3), from the laboratory at DLR TRON facility and at Airbus Robotic laboratory, from SurRender software high fidelity image simulator using Model Capture and from Generative Adversarial Networks. The use case definition included the selection of algorithms as benchmark: an AI-based pose estimation algorithm and a COTS dense optical flow algorithm were selected.  Eventually it is demonstrated that datasets produced with SurRender and selected laboratory facilities are adequate to train machine learning algorithms. 
\end{abstract}

\section{Introduction}
Project Training Datasets generation for machine learning: application to vision-based navigation is a GSTP study which ran from June 2022 to December 2023. It includes contributions from four teams, with a lead in Airbus Toulouse and co-lead in Airbus Bremen. Two subcontractors were involved: the French AI company Headmind and the DLR institute in Bremen.
The context is the introduction of machine learning in the field of Vision Based Navigation (VBN). VBN consists in utilising cameras as vision sensors for Guidance Navigation \& Control (GNC). Computer vision algorithms then extracts information from images to feed the navigation filter on the onboard computer. In recent years, general purpose applications and academic research has massively focused on deep learning. However there are obstacles for the adoption of machine learning algorithms for space. Challenges include the certification of these software for the stringent requirement of onboard software and the availability of standard development framework compatible with space processors - those are not considered in the framework of this project. Another challenge is the availability of input datasets to train the algorithms. This problem is the heart of the study. 

\section{Natural Scenario}
\subsection{Real and synthetic datasets}
We selected navigation camera images from Chang’e\,3 lander\cite{change3}. The input dataset includes images from the braking phase to touch down and they are formatted using the PDS standard. The trajectory is not directly available in the PDS header, rather lines-of-sights are provided. We inverted this information to retrieve an estimated trajectory.  
The scenario is then reproduced with synthetic simulations using the SurRender software, Airbus high performance image simulator\cite{brochard2018scientific,lebreton2021image,lebretoniac2022xxx}. Simulations include data fusion to merge multi-resolution terrain models, procedural details generation and metadata. There is no attempt to replicate exactly the real images: the ground truth trajectory has limited accuracy and the input are limited to Digital Elevation Models (DEM).  
The input model for the highest quality dataset is a low resolution global DEM from Chang’e-2 at 20m resolution \cite{rs12030526} merged with a high resolution Lunar Reconnaissance Orbiter (LRO) DEM at 5m resolution \cite{LRO-DEM}. The reflectance model is the Hapke BRDF (Bidirectional Reflectance Distribution Function), but the simulations do not include high resolution reflectance maps. Synthetic details (boulders, craters, Perlin noise) are added to a complementary dataset. 

\begin{figure}[h!]
    \centering
    \includegraphics[width=.95\columnwidth]{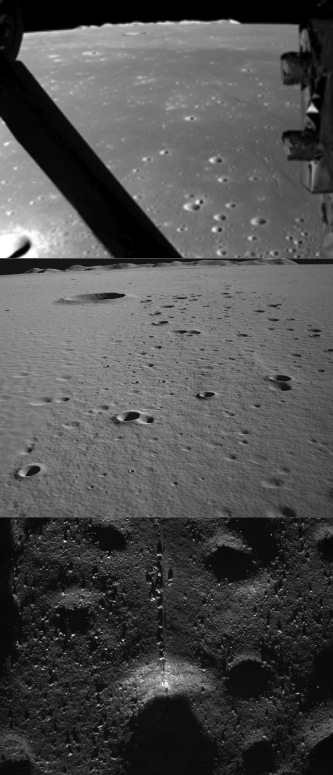}
	\caption{Extracts from Moon landing datasets. Top: real Chang’e-3 image. Middle: SurRender simulation with fusion of low-res and high-res DEMs. Bottom: Image captured with the TRON facility.}
	\label{fig:moondatasets}
\end{figure}

\subsection{TRON dataset}
The inverted Chang’e-3 trajectory was converted to robot control law to reproduce the scenario at DLR TRON facility\cite{KRUGER2010265}. This time the DEM is not related to the real landing site, but the landing dynamics is identical. An existing terrain model mockup was used as the target. It was mainly hand-crafted and measured afterwards with a 3D laser scanner. In addition to existing elements, a camera with relevant detector and field of view was attached to the robotic arm. A light source provides good illumination conditions (although with non-parallel light). 

\subsection{Ground truth}
The definition of ground truth depends on the targeted VBN algorithm. For camera poses, the accuracy of metadata is always limited and it is inherently worse than the targeted performance of VBN algorithm because it stems from other, less precise sensors. Hazard detection could be an interesting test case, however, there are no labels associated with real images. 
Rather, we selected a hybrid motion estimation algorithm to serve as an AI-based VBN metric. The ground truth consists in precomputed pixel motions in image pairs.

\section{Man-made object scenario}
\subsection{ENVISAT Laboratory datasets}
The Space Robotics Laboratory in Airbus Bremen and a mock-up of ENVISAT are used to generate laboratory datasets. The starting point was based on existing elements: a KUKA industrial robot and an industrial camera utilising a CMV4000 detector. An illumination unit emulating the Sun is added with a dark chamber. An ENVISAT mock-up at scale 1:33 was manufactured with the goal to produce a realistically looking satellite with precise knowledge of the as-built-status. The mock-up comprising the central block and the SAR antenna was milled at CNC milling accuracy. Golden foil was added to some surfaces to mimic MLI (Multi-layer insulation).
The ground truth is measured with high accuracy from the camera images using Aruco markers, precise calibration of the mockup/markers relative positions with an optical laser-tracker and a pose estimation algorithm. Final dataset products were background subtracted and provided with either black backgrounds, or random backgrounds. 

\subsection{SurRender synthetic renderings }
The SurRender software was also used to render images of ENVISAT. The computer model of the mockup was converted to graphical format without adding reflectance model input. In a first iteration for the low quality simulations, renderings sampled 3D space uniformly without additional attempts to increase realism. This dataset is an intermediate product which serves as input for the GAN. In a second approach the mockup model was postprocessed to remove manufacturing artefacts (holes) and to add realistic textures and BRDFs using model capture. Several datasets are produced, with or without the Earth in the background. The images are physically rendered in raytracing with realistic reflexions, secondary illumination, self-occlusion, etc. 

\subsection{Model capture: proof-of-concept}
An innovative method has been developed to extract textures and optical properties from the real laboratory images of the mockup and inject them in the simulations. SurRender 9 implements new features to allow basic model capture. Images can be backprojected into texture space to capture texture details and meaningful frequencies (normal mapping). Images could be differentiated with respect to basic parameters, meaning that gradients for some model parameters can be computed.
Taking advantage of this functionality, we could adjust BRDF parameters by maximising the resemblance between a small selection of photographs of the mockup and simulations. The output is a high definition ENVISAT model with visually realistic MLI models and reflectance properties.

\begin{figure}[h!]
    \centering
    \includegraphics[width=.95\columnwidth]{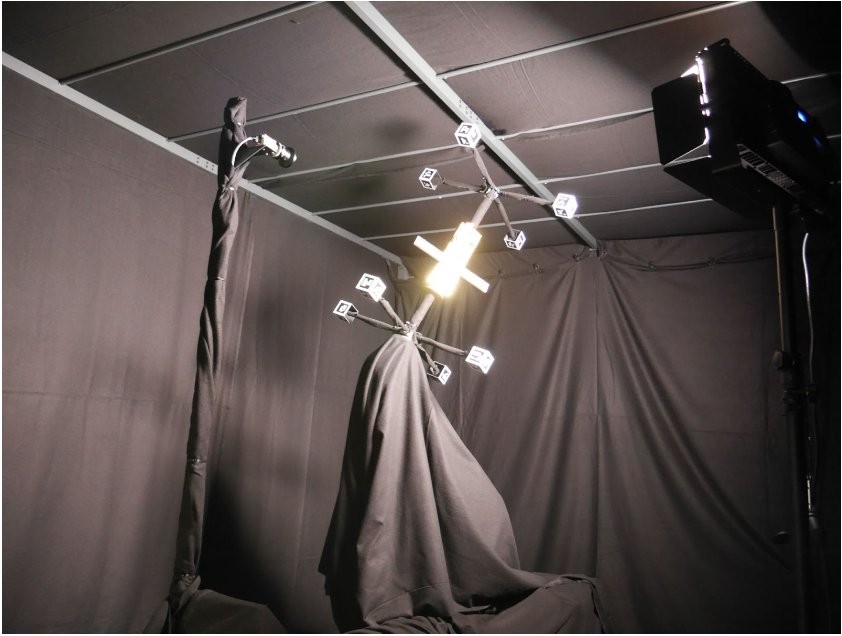}
	\caption{Real images of a mock-up of satellite ENVISAT were acquired inside a dark chamber at the Bremen robotic facility (right). This photograph shows the robotic arm (covered), the satellite mock-up and the illumination unit.}
	\label{fig:roboticlab}
\end{figure}

\begin{figure*}[h]
    \centering
    \includegraphics[width=.95\textwidth]{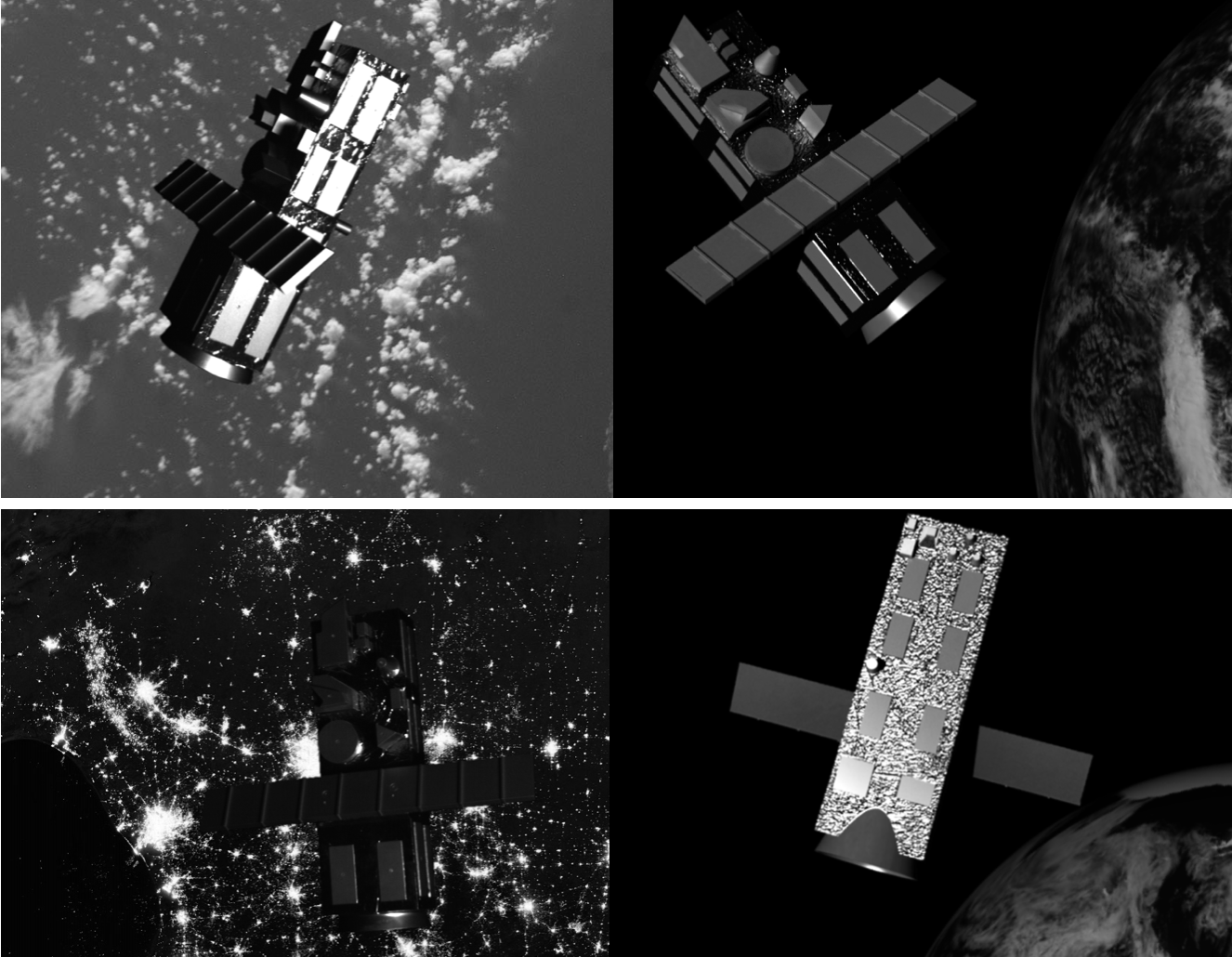}
	\caption{Examples of ENVISAT image datasets (cropped) generated either in the robotic laboratory (left) or with SurRender high fidelity simulations (right).}
	\label{fig:envisat}
\end{figure*}

\section{Generative Adversarial Network}
GANs are generative models used for Image-to-Image translation, such as transforming a source domain into a target domain\cite{GAN-review}. Our task involved converting low/medium definition SurRender simulations into images resembling real datasets. We developed a workflow to ensure ground truth accuracy, using both images and metadata. Training GANs is resource-intensive, taking days or weeks per learning iteration on a GPU.

For input domain A, we used medium definition synthetic images from Chang’e-3 for the natural scenario and low definition SurRender Envisat dataset for the man-made scenario. For target domain B, we used real images from the Chang’e-3 sequence and real images of the mockup from the Bremen robotic facility. We evaluated different GAN architectures and found that CUT (“Contrastive learning for unpaired image-to-image translation”) \cite{park2020contrastive} performed best, as it does not require paired datasets.

A major issue identified was the temporal incoherency in sequences, such as parasitic motions on the Lunar surface. To address this, we added metadata with geometrical constraints. Although the final results met high-level requirements by adding textures to low definition simulations and preserving ground truth at first order, some artifacts remained, such as aliasing, ghosts, and parasitic motions.

Future improvements could involve incorporating depth maps, pixel line of sight, or optical flow representations as metadata, and optimizing super-resolution or using full resolution inputs. While GAN technology has been surpassed by Diffusion Models, the general conclusion is that generative AI has not yet reached industrial readiness for VBN applications in space.

\begin{table*}[h!]\renewcommand{\arraystretch}{1.2} 
\begin{center}
\begin{tabular}{|l|l|l|l|l|} 
\hline\hline 
Scenario & Dataset & Type & Description & Number of images \\
\hline\hline 
Man-Made & MAN-DATA-S1 & Synthetic & Envisat HD & 16000 \\
\hline\hline 
Man-Made & MAN-DATA-S2 & Synthetic & Envisat HD - random backgrounds & 16000  \\
\hline\hline 
Man-Made & MAN-DATA-S5 & Synthetic & Envisat LD & 13875  \\
\hline\hline 
Man-Made & MAN-DATA-G1 & Synt. GAN & GAN Envisat & 13875  \\\hline\hline 
Man-Made & MAN-DATA-L1 & Laboratory & Laboratory Envisat & 16000  \\
\hline\hline 
Man-Made & MAN-DATA-L2 & Laboratory & Laboratory Envisat Background & 16000 \\
\hline\hline 
Natural & NAT-DATA-R1 & Real & Chang’e 3 Navcam & 3655  \\
\hline\hline 
Natural & NAT-DATA-L1 & Laboratory & TRON Testbed - Chang'e 3 & 3658  \\
\hline\hline 
Natural & NAT-DATA-L2 & Laboratory & TRON Testbed - Random pairs & 7238  \\
\hline\hline 
Natural & NAT-DATA-S1 & Synthetic & MD Chang'e 3 & 3655  \\
\hline\hline 
Natural & NAT-DATA-S2 & Synthetic & HD Chang'e 3 & 3655  \\
\hline\hline Natural & NAT-DATA-S3 & Synthetic & HD + procedural details Chang'e 3 & 3655  \\
\hline\hline Natural & NAT-DATA-G1 & Synt. GAN & MD + GAN Chang'e 3 & 1837  \\
\hline\hline Natural & NAT-DATA-S5 & Synthetic & HD, 3 simulated trajectories & 6661+5591+3736  \\
\hline\hline 
\end{tabular}
\caption{Overview of datasets produced during the project. LD: Low Definition. MD: Medium Definition, HD: High definition. The definition refers to the image quality for synthetic simulations rather than image resolution.}
\label{datasetoverview}
\end{center} \end{table*}

\section{Results}
\subsection{Man-made objects benchmark}
The pose estimation algorithm is a hybrid between an AI-based algorithm - a Convolutional Neural Network (CNN) used for the key-point detection - and a traditional algorithm - Perspective-n-Point (PnP)\cite{poseestimation}. Several metrics were exploited to quantify algorithm performance depending on the training dataset (Mean-Squared-Error between predicted and ground truth heatmaps, Keypoint Location Metric, Pose Estimation Metric).

The ENVISAT mock-up is visible from different directions, introducing the need for the neural network to detect key-points on the satellite from different sides, and the MLI fold introduced some harsh illumination conditions. 
This benchmark conclusion was that it is possible to train a network with synthetic datasets and apply it to laboratory datasets. Different compositions of laboratory datasets were tested, including raw data with markers, masked images without background and images with random earth images as background. The results show that occluded key-points are challenging to recognize for the neural network but that the detected key-points on the satellite front are sufficient for an accurate pose estimation.

The training of the neural network using the laboratory images and the test with laboratory images showed a higher accuracy than the training with synthetic images. This was expected when the same dataset is used for training and test. It was also expected that the mix of synthetic and laboratory images improves the performance compared to purely synthetic images.

\subsection{Moon landing scenario benchmark}
The optimization/regularisation step of a dense optical flow algorithm is a computationally expensive task which can be optimised by a CNN building block. We considered several algorithms available off-the-shelf and selected RAFT ("Recurrent All-Pairs Field Transforms")\cite{RAFT}, a new deep network architecture achieving state-of-the-art performances. It is important to highlight that the goal is not to develop the most efficient optical flow algorithm, which would require extensive data engineering, algorithm tuning and sizing, hyperparameters optimization and is out of the scope of this study. Pretrained models are available for reference, the objective of the benchmark is to train RAFT on the different datasets and analyse its performance. To quantify this performance, the applied metrics are the optical flow end-point error (EPE) aggregated in a single scale and visual inspection.
Overall, it is found that models trained on SurRender and TRON datasets reproducing the Chang’e-3 trajectory perform significantly better on the real data (real Chang’e-3 images) than the pretrained model. This shows that these datasets are suitable for training. The GAN dataset performed the worst but given the density of artefacts and lack of temporal coherency it was expected.

\begin{figure*}[h!]
    \centering
 \includegraphics[width=.45\textwidth]{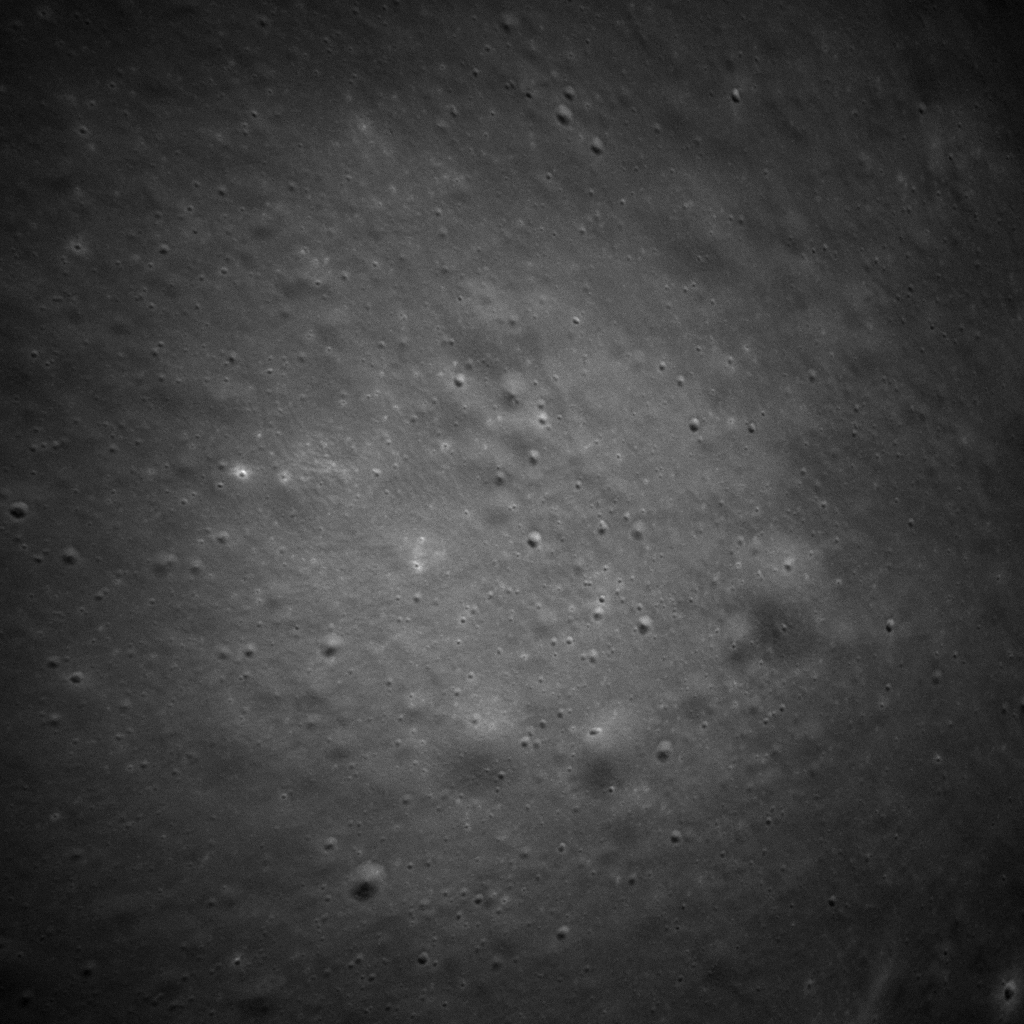}
    \includegraphics[width=.45\textwidth]{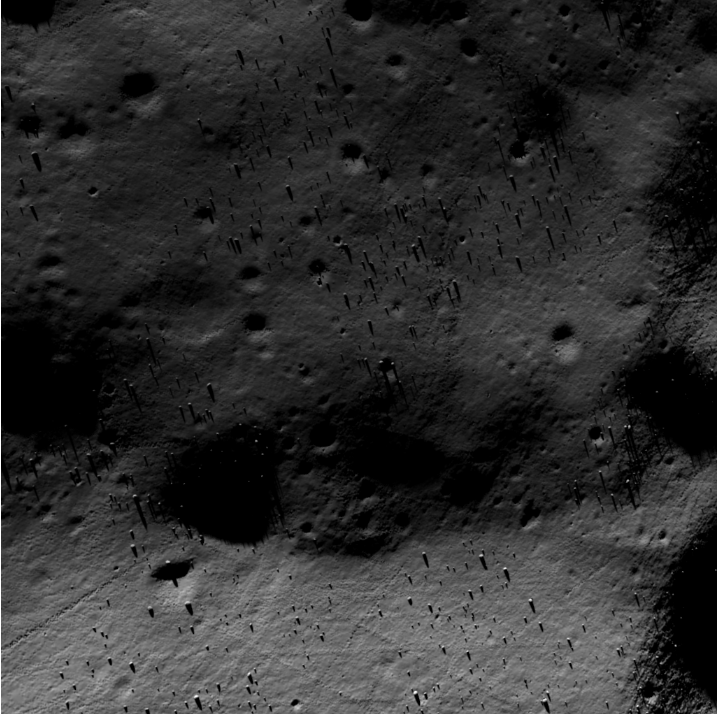}
        \caption{SurRender renderings of the Moon surface using high resolution reflectance maps in addition to DEMs. In the right image, a distribution of boulders is projected on the terrain.}
        \label{fig:surrender_reflectance}
\end{figure*}

\section{Discussion}
These results answer the main question raised by the study: datasets generated with synthetic simulations (SurRender) and on robotic benches can be exploited to train an AI, which is then able to generalise to real datasets. For the optical flow, the training yields much better performance than pre-trained models. 
The study can serve as reference to validate future datasets produced with a similar process. The toolset developed during the study has reached high maturity and it could be exploited to produce new datasets. The delivered datasets are now ready to be shared under ESA Community License. 
As a way forward, several topics could be explored more, for example: enhancing the synthetic datasets with additional metadata (labels, kinematic) and a diversity of trajectories, generalising the datasets with a range of parameters for input models, revisiting generative AI techniques applying lesson-learnt for a more robust workflow, or experimenting model capture on Moon datasets. \\

Acknowledgment: This study was carried our by Airbus for the European Space Agency under ESA Contract No. 4000137956/22/NL/CRS. We are thankful to Emma Erre from Headmind partner for her involvment in the GAN task and to Hans Krüger from DLR for the production of the TRON dataset. We thank the LROC team / NASA and the Chang'E-2 team / CNSA for providing part of the data that made this study possible. \\

Preprint from the proceedings of ESA SPAICE conference 2024.  

\printbibliography
\addcontentsline{toc}{section}{References}

\end{document}